\documentclass[10pt]{article}
\usepackage[utf8]{inputenc}
\usepackage[T1]{fontenc}
\usepackage{amsmath}
\usepackage{amsthm}
\usepackage{amsfonts}
\usepackage{amssymb}
\usepackage{makeidx}
\usepackage{graphicx}
\usepackage{algorithm2e}
\usepackage{lmodern}
\usepackage{mathtools}
\usepackage{url}
\usepackage{listings}

\usepackage{xcolor}
\usepackage[left=2cm,right=2cm,top=1cm,bottom=2cm]{geometry}
\usepackage{multirow}
\usepackage{caption}
\usepackage{subcaption}
\usepackage{tabularx} 
\usepackage{booktabs}

\author{ Prudence Djagba\thanks{ Michigan State University
		\href{mailto:arame.sow@aims-senegal.org}{ djagbapr@msu.edu}}
\and 
Chimezie A. Odinakachukwu   \thanks{AIMS Senegal. 
		\href{chimezie.a.odinakachukwu@aims-senegal.org}
{chimezie.a.odinakachukwu@aims-senegal.org}} 
    }
\title{ Assessing the Capabilities and Limitations of FinGPT Model in Financial NLP Applications}

\usepackage{hyperref}
\usepackage{cleveref}

\theoremstyle{remark}

\theoremstyle{definition}

\date{}
\begin{document}
\maketitle	
\begin{abstract} 

This work evaluates FinGPT, a financial domain-specific language model, across six key natural language processing (NLP) tasks: Sentiment Analysis, Text Classification, Named Entity Recognition, Financial Question Answering, Text Summarization, and Stock Movement Prediction. The evaluation uses finance-specific datasets to assess FinGPT’s capabilities and limitations in real-world financial applications. The results show that FinGPT performs strongly in classification tasks such as sentiment analysis and headline categorization, often achieving results comparable to GPT-4. However, its performance is significantly lower in tasks that involve reasoning and generation, such as financial question answering and summarization. Comparisons with GPT-4 and human benchmarks highlight notable performance gaps, particularly in numerical accuracy and complex reasoning. Overall, the findings indicate that while FinGPT is effective for certain structured financial tasks, it is not yet a comprehensive solution. This research provides a useful benchmark for future research and underscores the need for architectural improvements and domain-specific optimization in financial language models.
\end{abstract}


\section{Introduction}
The financial industry has long been a pioneer in adopting cutting-edge technologies to enhance operational efficiency, accuracy, and strategic decision-making \cite{alabi2022ai}. With the exponential growth of structured and unstructured data, particularly from news feeds, earnings reports, disclosures, and social media, there is an increasing demand for intelligent systems capable of processing human language at scale \cite{chlouverakis2024artificial}. Initially, the industry relied on rule-based approaches and traditional statistical techniques such as bag-of-words and TF-IDF \cite{ozbayoglu2020deep}, which offered limited semantic understanding. As noted by Abubakar et al.\cite{abubakar2022sentiment}, these limitations triggered a shift toward machine learning and deep learning models that, while better at capturing patterns, still required substantial domain-specific feature engineering.

This landscape was significantly transformed with the introduction of transformer-based architectures, most notably the Generative Pre-trained Transformer (GPT) family \cite{brown2020language}. These models demonstrated the power of large-scale pretraining followed by task-specific fine-tuning, enabling generalization across diverse NLP tasks. Models such as GPT-3, GPT-4, BERT, and T5 have delivered state-of-the-art results in sentiment analysis, summarization, question answering, and named entity recognition \cite{devlin2019bert}. Beyond LLMs, the broader field of Generative AI (GAI)—including GANs, VAEs, and diffusion models—has found increasing relevance in finance, facilitating applications such as synthetic data generation, automated reporting, and scenario simulation \cite{shabsigh2023generative, rizzato2023generative}.

LLMs have emerged as essential tools in processing unstructured financial text, especially models fine-tuned on finance-specific corpora like FinBERT, BloombergGPT, and FinGPT \cite{araci2019finbert, wu2023bloomberggpt}. Their capabilities in capturing complex linguistic nuances and domain-specific terminology make them highly effective in financial NLP tasks. However, challenges persist due to the specialized jargon, numerical reasoning, and the high-stakes context inherent in financial text. As highlighted by Qian et al.\cite{qian2025fino1}, understanding and improving the capabilities of these models through fine-tuning and rigorous benchmarking remains crucial to their reliable integration in finance. This intersection of GAI and finance thus provides a rich domain for innovation and intelligent decision-support systems.

\subsection{Background}

Recent advances in large language models (LLMs), such as GPT-3, GPT-4, and their instruction-tuned variants, have revolutionized general-purpose natural language understanding. However, applying these models effectively to domain-specific tasks, particularly in finance, remains a significant challenge. Financial texts often contain specialized vocabulary, abbreviations, and implicit knowledge, which general models are not always optimized to handle.

While domain-specific models like \textbf{FinGPT} and \textbf{FinMA 7B} have emerged to address these gaps, their performance across a diverse set of financial NLP tasks remains uneven and often under-evaluated. Guo et al.~\cite{guo2023chatgpt} noted that many evaluations are narrow in scope, and Li et al.~\cite{li2023chatgpt} emphasized the lack of systematic benchmarks and evaluation strategies across key financial tasks such as sentiment analysis, named entity recognition, and financial question answering.

Furthermore, there has been limited comparative analysis between general-purpose models like GPT-4 and domain-tuned models like FinGPT. This lack of comparison leaves unanswered questions about performance trade-offs in terms of accuracy, interpretability, and resource efficiency—critical considerations when deploying models in real-world financial systems where retraining is expensive.

To address this gap, this research evaluates the performance of FinGPT on six core financial NLP tasks and compares it with GPT-4 and FinMA 7B:

\begin{table}[h]
\centering
\caption{Overview of the Six NLP Tasks and Their Corresponding Datasets}
\label{tab:six_nlp_tasks}
\scriptsize
\begin{tabular}{|l|p{7.5cm}|p{4.5cm}|}
\hline
\textbf{NLP Task} & \textbf{Dataset} & \textbf{Citation} \\
\hline
Sentiment Analysis (SA) & FLARE-FPB, FinQA-SA & \cite{flare-fpb}, \cite{finqa-sa} \\
\hline
Text Classification (TC) & FinGPT-Headline & \cite{fingpt-headline} \\
\hline
Named Entity Recognition (NER) & FinGPT-NER & \cite{fingpt-ner} \\
\hline
Financial Question Answering (QA) & ConvFinQA, FLARE-FinQA & \cite{convfinqa}, \cite{flare-finqa} \\
\hline
Stock Movement Prediction (SMP) & CIKM18, StockNet, BigData22 & \cite{cikm18}, \cite{stocknet}, \cite{bigdata22} \\
\hline
Text Summarization (Summ) & ECTSum & \cite{ectsum} \\
\hline
\end{tabular}
\end{table}

This research contributes a structured benchmark comparison and a practical methodology for evaluating large language models in financial NLP pipelines—balancing accuracy, efficiency, and domain alignment.

\subsection{Research Questions}
To guide this investigation, the following research questions were investigated:

\begin{itemize}
    \item \textbf{RQ1:} How well does FinGPT perform on six core financial NLP tasks using real-world finance-specific datasets?
    \item \textbf{RQ2:} How does FinGPT's performance compare with that of general-purpose models like GPT-4 and domain-tuned models like FinMA 7B on the same tasks?
    \item \textbf{RQ3:} What performance limitations emerge across tasks, and what insights can be drawn to guide future development of financial LLMs?
\end{itemize}

\section{Evolution of Generative AI and Financial LLMs}

The history of artificial intelligence (AI) dates back to the Dartmouth Summer Research Project (1956) \cite{wired_dartmouth_2012}, which envisioned that machine intelligence could replicate human learning. Since then, AI has undergone several milestones, particularly with the rise of generative models. Unlike discriminative models focused on classification, generative models aim to learn and recreate the data distribution, making them well-suited for text generation and structured reasoning.

Early innovations include Variational Autoencoders (VAEs) \cite{lee2024comprehensive} and Generative Adversarial Networks (GANs) \cite{goodfellow2016deep}, which laid the foundation for modern generative systems. The introduction of Transformer architectures by Vaswani et al. in 2017 \cite{vaswani2017attention} marked a pivotal shift in deep learning, enabling scalable and context-aware sequence modeling. This breakthrough gave rise to the Generative Pre-trained Transformer (GPT) series—GPT-2, GPT-3, and GPT-4—which demonstrated exceptional performance across natural language processing (NLP) tasks including summarization, question answering, and translation \cite{brown2020language, yenduri2024gpt}.

In finance, the shift from bag-of-words to transformer-based LLMs enabled models to capture domain-specific context more effectively. Pre-trained models like BERT \cite{devlin2019bert} led to specialized variants such as FinBERT \cite{araci2019finbert} and FinBERT-21 \cite{huang2023finbert}, improving sentiment analysis and entity recognition in financial texts. Proprietary models like BloombergGPT \cite{wu2023bloomberggpt} trained on hybrid corpora expanded task coverage but remained inaccessible to most users due to licensing and compute barriers.

\begin{figure}[h]
    \centering
    \includegraphics[width=1\linewidth]{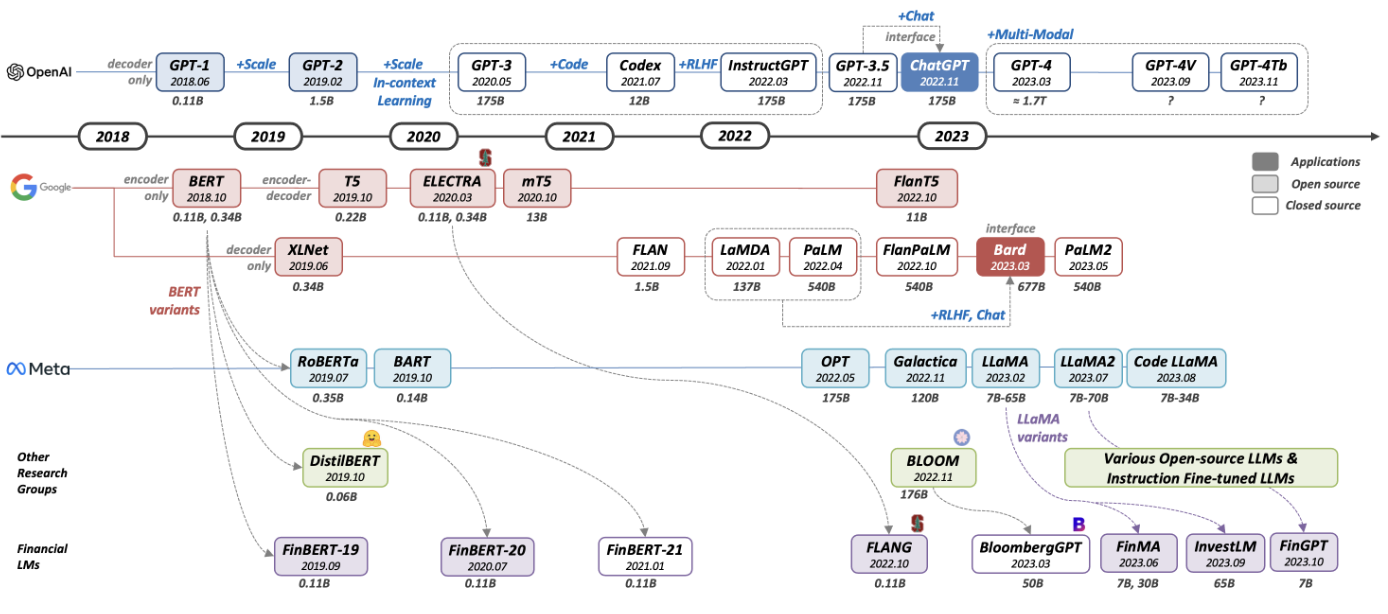}
    \caption{Timeline showing the evolution of selected PLM/LLM releases from the general domain to the financial domain \cite{lee2024survey}.}
    \label{fig: Timeline}
\end{figure}

\begin{figure}[h!]
    \centering
    \includegraphics[width=1\linewidth]{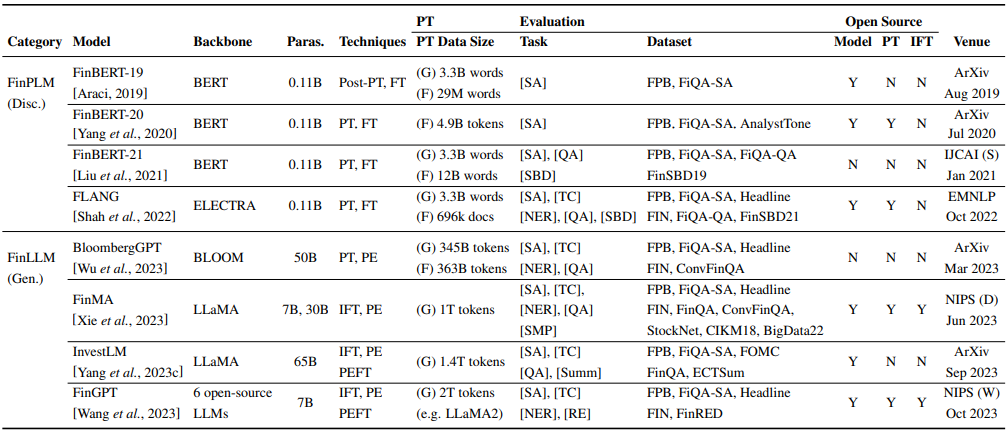}
    \caption{Summary of FinPLMs and FinLLMs. Abbreviations include Paras. = Parameter Size, PT = Pretraining, FT = Fine-Tuning, SA = Sentiment Analysis, QA = Question Answering, SMP = Stock Movement Prediction, Summ = Summarization, and others \cite{liu2023fingpt}.}
    \label{fig:summary_table}
\end{figure}

Open-source LLMs such as the LLaMA series \cite{touvron2023llama} and FinGPT \cite{liu2023fingpt} offer scalable alternatives, often leveraging parameter-efficient fine-tuning (PEFT) and real-time training. Similarly, FinMA \cite{xie2023pixiu} and InvestLM \cite{yang2023investlm} are optimized for financial tasks like market analysis and regulatory summarization. Figure~\ref{fig: Timeline} and Figure~\ref{fig:summary_table} illustrate the evolution from general to domain-specific LLMs in finance.

\subsection{Challenges in Financial AI Deployment}

Deploying LLMs in finance introduces unique challenges. Model transparency, data privacy, and regulatory compliance are critical concerns. Black-box behaviors and hallucinations hinder trust, while limited labeled data restricts training quality \cite{rankovic2023artificial, lewis2020retrieval}.

Retrieval-Augmented Generation (RAG) offers a promising solution by allowing models to access private databases during inference without retraining \cite{lewis2020retrieval}. Meanwhile, evaluation metrics like accuracy and F1 often fall short in assessing financial task performance, prompting a shift toward expert-in-the-loop assessment.

Infrastructure constraints also limit adoption, especially due to high computational costs and memory demands. Open-access models such as FinGPT offer a cost-effective alternative but still lag behind GPT-4 in complex tasks. Models like FinGPT-13B remain underutilized due to resource limits.

\newpage

\subsection{Enabling Responsible AI in Finance}

To foster responsible AI use, federated learning (FL) \cite{dhanawat2024enhancing} enables collaborative model training across institutions while preserving data privacy. Explainable AI (XAI), particularly SHAP values, improves model interpretability for sensitive decisions like fraud detection and credit scoring.

Lastly, upskilling professionals and bridging the AI-literacy gap are essential. Industry-academic partnerships and targeted curriculum reforms are crucial to ensure long-term adaptability and trust \cite{dalwai2022investigation, aldemir2024ai, iwuanyanwu2023analyzing}.

This shift from general-purpose to specialized financial LLMs highlights the growing importance of transparency, adaptability, and scalability in real-world financial NLP applications.

\section{Methods} 

\subsection{Overview of FinGPT Framework}

\begin{figure}[h!]
    \centering
    \includegraphics[width=\linewidth]{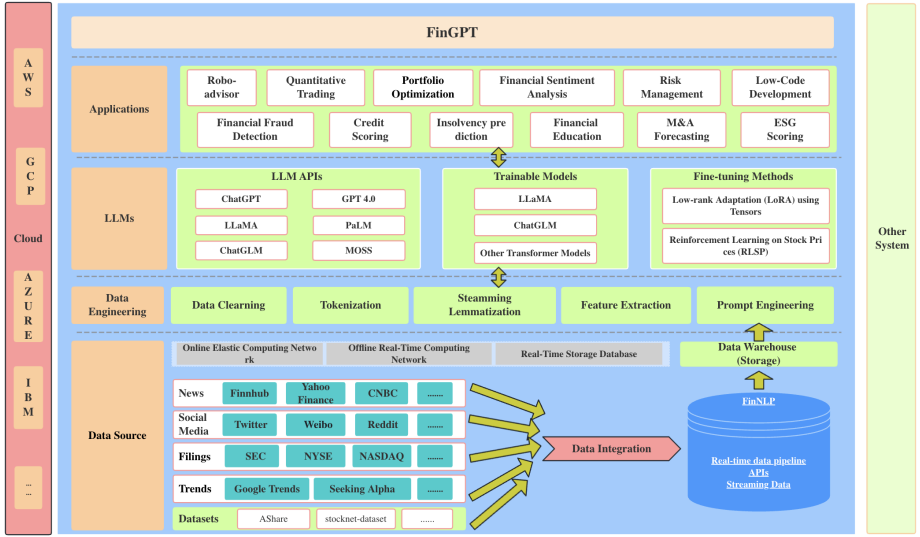}
    \caption{FinGPT system architecture highlighting the end-to-end data and modeling pipeline \cite{yang2023fingpt}.}
    \label{fig:fingpt_framework}
\end{figure}

FinGPT is an open-source framework designed to enable the application of Large Language Models (LLMs) in finance. It addresses challenges such as market volatility, heterogeneous data sources, and real-time text processing. As illustrated in Figure~\ref{fig:fingpt_framework}, the FinGPT architecture consists of four modular layers:

\begin{itemize}
    \item \textbf{Data Source:} Collects structured and unstructured data from financial APIs, social media, news, and filings.
    \item \textbf{Data Engineering:} Handles cleaning, labeling, and streaming of data.
    \item \textbf{LLMs:} Utilizes transformer-based architectures (e.g., LLaMA) fine-tuned for finance.
    \item \textbf{Applications:} Deploys models to various tasks such as sentiment analysis, question answering, and stock movement prediction.
\end{itemize}

\subsection{LLaMA2: Model Foundation for FinGPT}

FinGPT is based on the LLaMA2 (Large Language Model Meta AI) architecture, a decoder-only transformer introduced by Touvron et al. \cite{touvron2023llama}. The LLaMA2 model inherits the Transformer decoder architecture originally proposed by Vaswani et al. \cite{vaswani2017attention}, optimized for autoregressive language modeling.

\subsubsection{Transformer Decoder Architecture}

The model comprises $N$ stacked decoder blocks, each consisting of:

\begin{enumerate}
    \item Multi-Head Self-Attention (MHSA)
    \item Feed-Forward Networks (FFN)
\end{enumerate}

For an input sequence $X = (x_1, x_2, \dots, x_T)$ with embeddings $x_i \in \mathbb{R}^d$, the model computes token-wise representations through successive MHSA and FFN layers, incorporating residual connections and normalization.

\subsubsection{Multi-Head Self-Attention}

Each attention head is computed as:
\[
\text{Attention}(Q, K, V) = \text{softmax}\left(\frac{QK^\top}{\sqrt{d_k}}\right)V
\]
where $Q = XW^Q$, $K = XW^K$, $V = XW^V$ are projections with learnable matrices $W^Q$, $W^K$, and $W^V \in \mathbb{R}^{d \times d_k}$.

The multi-head extension is given by:
\[
\text{MHSA}(X) = \text{Concat}(head_1, \dots, head_h)W^O
\]

\subsubsection{Rotary Positional Embedding (RoPE)}

LLaMA replaces absolute positional encodings with Rotary Positional Embeddings (RoPE), introducing position-dependent rotations to maintain relative token information \cite{su2024roformer}. This enhances generalization across varying sequence lengths.

\subsubsection{Feed-Forward Networks}

Each token's representation is passed through a two-layer FFN:
\[
\text{FFN}(x) = W_2 \cdot \text{GELU}(W_1 x + b_1) + b_2
\]

\subsubsection{Normalization and Residual Connections}

Layer outputs are stabilized via pre-layer normalization and residual connections:
\[
x' = x + \text{MHSA}(\text{LayerNorm}(x)), \quad x'' = x' + \text{FFN}(\text{LayerNorm}(x'))
\]

\subsubsection{Training Objective}

The model is trained using next-token prediction by minimizing the negative log-likelihood:
\[
\mathcal{L} = -\sum_{t=1}^{T} \log P(x_t \mid x_{<t}; \theta)
\]

\subsection{Adaptation in FinGPT}

FinGPT applies domain-specific adaptation to LLaMA through fine-tuning on financial corpora. While the underlying transformer structure remains intact, adaptations include:
\begin{itemize}
    \item Task-aligned instruction tuning (e.g., classification, QA)
    \item Financial dataset pretraining and finetuning
    \item Lightweight model variants for lower resource environments
\end{itemize}

These adaptations allow FinGPT to address complex financial NLP tasks with improved accuracy and relevance.

\section{Evaluation}

\subsection{Task Overview}

To evaluate FinGPT's applicability across financial natural language processing (NLP) tasks, we designed a standardized evaluation pipeline involving six key tasks: sentiment analysis, text classification, named entity recognition, question answering, stock movement prediction, and text summarization. The methodology includes dataset selection, preprocessing, model configuration, inference, and metric-based evaluation.

\subsection{Sentiment Analysis}

\subsubsection{Datasets}

Two open-source datasets were selected:

\begin{itemize}
    \item \textbf{FLARE-FPB}~\cite{flarefpb2024}: Contains labeled financial texts (positive, neutral, negative); 970 samples from the test split were used.
    \item \textbf{FLARE-FIQASA}~\cite{finqa-sa}: Comprises 235 financial QA-style texts; labels were extracted from the \texttt{answer} field.
\end{itemize}

\subsubsection{Preprocessing}

The preprocessing stage involved formatting each sample as an instruction-style prompt, where the model was explicitly guided to identify sentiment from financial text. All inputs were lowercased and normalized to ensure consistency across label representations. Tokenization was performed using the \texttt{AutoTokenizer} from the Hugging Face Transformers library, with padding aligned to the end-of-sequence (EOS) token to ensure compatibility with the decoder-only architecture. Notably, no aggressive text cleaning was applied in order to preserve domain-specific entities such as stock tickers (e.g., \texttt{\$AAPL}) and financial indicators.

\subsubsection{Model Configuration}

The experiments employed the \texttt{NousResearch/Llama-2-13b-hf} model as the base architecture. This model was extended using a Low-Rank Adaptation (LoRA) module, specifically the adapter \texttt{oliverwang15/FinGPT\_v33\_...}, tailored for sentiment classification in finance. To optimize memory usage and enable efficient inference, the model was loaded in 8-bit precision with weights stored in \texttt{float16}. The model was set to evaluation mode using \texttt{model.eval()} to deactivate training layers and ensure deterministic behavior during generation.

\subsubsection{Inference}

Inference was performed in batches of four for the FLARE-FIQASA dataset, while a single-batch mode was used for FLARE-FPB due to differing sequence lengths. Text generation was executed using greedy decoding (\texttt{do\_sample=False}) with a maximum output length of 32 tokens for FLARE-FPB and 512 tokens for FLARE-FIQASA to accommodate longer responses. The generated outputs were subsequently normalized and mapped to the expected sentiment categories to facilitate evaluation.

\subsection{Text Classification}

\subsubsection{Dataset}

The \texttt{FinGPT Headline Classification} dataset~\cite{fingpt-headline} was used, containing financial headlines labeled with binary sentiment (\texttt{yes}/\texttt{no}).

\subsubsection{Preprocessing}

Each input headline was embedded into an instruction-style template to guide the model's classification task. The prompt format followed the structure:

\begin{lstlisting}
[INST] Classify the sentiment of the following financial headline: <HEADLINE> [/INST]
\end{lstlisting}

Tokenization was handled using the \texttt{AutoTokenizer} associated with the LLaMA-2 model. Left-padding was applied, and special attention was given to align special tokens with the EOS marker. To ensure consistent label evaluation, synonymous expressions such as “absolutely” or “nope” were normalized and mapped to binary sentiment labels. Ambiguous cases labeled as \texttt{maybe}, or outputs that could not be confidently categorized, were excluded from final evaluation to avoid metric distortion.

\subsubsection{Model Setup}

The classification model used \texttt{meta-llama/Llama-2-7b-hf} as the base architecture. To adapt it for financial sentiment classification without full fine-tuning, a Low-Rank Adaptation (LoRA) adapter from the FinGPT project
\texttt{FinGPT/fingpt-mt\_llama2-7b\_lora}—was applied using the \texttt{peft} (Parameter-Efficient Fine-Tuning) library.

\subsubsection{Inference and Prediction}

During inference, the model generated sentiment predictions using the \texttt{generate()} function, constrained to a maximum of 10 tokens to enforce concise outputs. The generated outputs were decoded and passed through a normalization function that mapped the text to standardized sentiment categories (\texttt{yes}, \texttt{no}, or \texttt{unknown}) based on keyword matching:

\begin{lstlisting}
def extract_label(output_text):
    output_text = output_text.lower()
    if "yes" in output_text:
        return "yes"
    elif "no" in output_text:
        return "no"
    else:
        return "unknown"
\end{lstlisting}

\subsubsection{Evaluation Metrics}

Model performance was assessed using standard classification metrics, including precision, recall, and F1-score, calculated over the \texttt{yes} and \texttt{no} classes. Any outputs categorized as \texttt{unknown}—due to lack of recognizable sentiment indicators –were excluded from the score aggregation to maintain the reliability of the evaluation.

\subsection{Named Entity Recognition (NER)}

\subsubsection{Dataset Description}

The Named Entity Recognition (NER) task was evaluated using the publicly available \texttt{FinGPT/fingpt-ner} dataset from Hugging Face (\url{https://huggingface.co/datasets/FinGPT/fingpt-ner}). This dataset contains financial-domain sentences annotated with entities from three classes: \texttt{person}, \texttt{organization}, and \texttt{location}. Each sample comprises a sentence and an expected output string in the format ``Entity is a [type]''. The test set, consisting of 98 examples, was used as-is for zero-shot or few-shot evaluation without additional annotation or manual filtering.

\subsubsection{Preprocessing}

Each sentence was tokenized using the LLaMA-2 tokenizer (\texttt{meta-llama/Llama-2-7b-hf}) with left-padding and a maximum token length of 512 to suit decoder-style generation. A consistent instruction-style prompt was applied to all inputs:

\begin{lstlisting}
Instruction: Please extract entities and their types from the input sentence,
entity types should be chosen from {person/organization/location}.
Input: <sentence>
Answer:
\end{lstlisting}

The generated outputs were post-processed to align with BIO tagging standards. Entities were extracted and mapped to token spans using the B-\{TYPE\}, I-\{TYPE\}, and O schema. Samples with token-output mismatches were excluded, although none were discarded during this experiment.

\subsubsection{Model Configuration}

The experiment used the \texttt{meta-llama/Llama-2-7b-hf} model as the base, with a parameter-efficient LoRA adapter from the FinGPT project (\texttt{FinGPT/fingpt-mt\_llama2-7b\_lora}) for financial-domain specialization. The model and adapter were integrated using the \texttt{transformers} and \texttt{peft} libraries. Evaluation was conducted using \texttt{float16} precision and automatic GPU allocation via \texttt{device\_map="auto"}.

\subsubsection{Inference Procedure}

The model was prompted using the instruction-based format and decoded using greedy decoding (\texttt{do\_sample=False}). Extensive parameter tuning was required to ensure performance:

\begin{itemize}
    \item \textbf{Maximum Length:} Set to 500 to balance output coherence and memory usage.
    \item \textbf{Max New Tokens:} Reduced from 64 to 34, which improved macro F1 from 38\% to approximately 69\%, minimizing hallucination.
    \item \textbf{Batch Size:} Set to 1 to accommodate hardware constraints and prevent instability during decoding.
\end{itemize}

These adjustments underscore the importance of careful generation parameter calibration when using autoregressive models like LLaMA for structured tasks such as NER. Despite architectural limitations, the model achieved robust entity extraction when appropriately tuned.

\subsection{Financial Question Answering}
\label{sec:QA}

\subsubsection{Dataset Description}

To assess FinGPT’s capability in financial question answering, we evaluated its performance on two benchmark datasets: \textbf{ConvFinQA} and \textbf{FLARE-FinQA}.

\textbf{ConvFinQA} is designed for multi-step numerical reasoning within a conversational context. It was accessed from Hugging Face at \url{https://huggingface.co/datasets/FinGPT/fingpt-convfinqa}, and the first 200 examples from the \texttt{test} split were selected. Each sample comprises a question and a numeric answer, requiring arithmetic reasoning and context comprehension.

\textbf{FLARE-FinQA} was obtained from \url{https://huggingface.co/datasets/ChanceFocus/flare-finqa}, and the first 50 test samples were used. It contains natural language queries with corresponding numeric or binary (``yes''/``no'') answers, testing both factual recall and logical reasoning in financial contexts.

\subsubsection{Preprocessing}

Preprocessing included prompt formatting, tokenization, answer extraction, and filtering. Both datasets were converted into a consistent prompt format to elicit concise numerical responses:

\begin{lstlisting}
Please answer the given financial question based on the context.
Question: {question}
Answer:
\end{lstlisting}

The true answers were extracted from each dataset’s ground truth field. Tokenization was applied using the LLaMA-2 tokenizer with padding and truncation, using a maximum sequence length of 768 for FLARE-FinQA and 1012 for ConvFinQA. Batch sizes were set to 4 and 8 respectively.

Model outputs were cleaned with regular expressions to isolate numeric values. Only examples where both prediction and ground truth were valid numbers were retained for evaluation, ensuring fairness and consistency in scoring.

\subsubsection{Model Configuration}

The experiments utilized a parameter-efficient fine-tuning (PEFT) setup based on the \texttt{meta-llama/Llama-2-7b-hf} model. A domain-specific LoRA adapter from the FinGPT project (\texttt{FinGPT/fingpt-mt\_llama2-7b\_lora}) was integrated using the \texttt{peft} library.

The model was loaded with \texttt{float16} precision and assigned automatically to available GPUs via \texttt{device\_map="auto"} to optimize runtime performance.

\subsubsection{Inference Strategy}

Inference was conducted using beam search with 2 beams and a temperature of 0.7. The maximum generation length was capped at 34 tokens to constrain output length and reduce hallucination. The model was executed within a \texttt{torch.no\_grad()} context to minimize memory overhead during evaluation.

After generation, the predicted outputs were parsed to extract numerical values for comparison with ground truth. For FLARE-FinQA, if multiple numbers were present, the one with the smallest absolute difference from the true value was used as the final prediction.

This setup allowed us to benchmark FinGPT’s ability to perform structured, quantitative financial reasoning under realistic evaluation conditions.

\subsection{Stock Movement Prediction}

\subsubsection{Dataset Description}

To evaluate FinGPT’s performance on the Stock Movement Prediction (SMP) task, we employed three publicly available datasets accessed via Hugging Face. These datasets contain aligned financial news content and corresponding stock movement labels suitable for classification tasks:

\begin{itemize}
    \item \textbf{CIKM18 (flare-ECTSum)} \cite{cikm18} \\
    \textit{Source:} \url{https://huggingface.co/datasets/ChanceFocus/flare-ectsum} \\
    A benchmark dataset from the CIKM 2018 Financial Opinion Mining Challenge. It includes news summaries annotated with subsequent stock movement labels (up or down), providing insight into event-driven market response.

    \item \textbf{StockNet (flare-SM-ACL)} \cite{stocknet} \\
    \textit{Source:} \url{https://huggingface.co/datasets/ChanceFocus/flare-sm-acl} \\
    Comprises sentiment-enriched financial news texts and corresponding price direction labels. It is widely adopted for evaluating sentiment-based stock forecasting models.

    \item \textbf{BigData22 (flare-SM-BigData)} \cite{bigdata22} \\
    \textit{Source:} \url{https://huggingface.co/datasets/TheFinAI/flare-sm-bigdata} \\
    A large-scale dataset integrating diverse financial sources such as news articles and social media with stock price annotations. It supports broader-scale evaluation under high-volume scenarios.
\end{itemize}

All datasets were loaded using the \texttt{datasets} library with the \texttt{split="test"} configuration. Labels were provided in binary or ternary format (\texttt{up}, \texttt{down}, \texttt{stable}), enabling a realistic multi-class prediction scenario.

\subsubsection{Preprocessing}

Standard preprocessing was performed to ensure consistency and compatibility across all datasets:

\begin{itemize}
    \item \textbf{Text Cleaning:} Raw texts were cleaned by removing URLs, redundant whitespace, and special symbols. All inputs were lowercased for uniformity.

    \item \textbf{Date Alignment:} Stock movement labels were matched to news reports based on timestamps. Misaligned or incomplete entries were excluded.

    \item \textbf{Label Encoding:} Movement directions were mapped to numeric values: \texttt{up} $\rightarrow$ 1, \texttt{down} $\rightarrow$ 0, and \texttt{stable} handled conditionally based on task formulation.

    \item \textbf{Tokenization:} Cleaned text was tokenized using the LLaMA-2 tokenizer with padding and truncation applied to respect maximum sequence lengths.

    \item \textbf{Batching:} Preprocessed inputs were grouped using PyTorch’s \texttt{DataLoader} to enable efficient GPU-accelerated inference.
\end{itemize}

\subsubsection{Model Configuration}

The model used for this task was \textbf{FinGPT-Forecaster}, a parameter-efficient fine-tuned variant of \texttt{LLaMA-2-7B-chat}:

\begin{itemize}
    \item \textbf{Base Model:} \texttt{meta-llama/Llama-2-7b-chat-hf}, a decoder-only language model.

    \item \textbf{LoRA Adapter:} \texttt{FinGPT/fingpt-forecaster\_dow30\_llama2-7b\_lora}, a fine-tuned checkpoint tailored to Dow Jones-related stock prediction.

    \item \textbf{Tokenizer:} The matching LLaMA-2 tokenizer was used to prepare inputs.

    \item \textbf{Hardware and Precision:} The model was loaded using \texttt{float16} precision and automatically deployed to GPU using \texttt{device\_map="auto"} for optimized execution.

    \item \textbf{Libraries:} Hugging Face’s \texttt{transformers} and \texttt{peft} libraries were used for model initialization and adapter integration.
\end{itemize}

\subsubsection{Inference Strategy}

The inference procedure followed a structured and reproducible pipeline:

\begin{itemize}
    \item \textbf{Prompt Construction:} Each news summary or article was embedded into a structured prompt suitable for the LLaMA-2-chat model.

    \item \textbf{Generation:} The model generated textual outputs (e.g., ``up'' or ``down'') using greedy decoding or controlled generation parameters.

    \item \textbf{Prediction Extraction:} Output strings were parsed with regular expressions to extract the predicted class.

    \item \textbf{Batch Execution:} Predictions were performed in mini-batches using PyTorch, leveraging GPU acceleration.

    \item \textbf{Postprocessing:} Textual predictions were mapped back to numeric labels to facilitate quantitative evaluation.
\end{itemize}

This pipeline ensured compatibility across datasets, efficiency in execution, and consistency in prediction output, forming a solid foundation for analyzing FinGPT’s forecasting capabilities on financial time series events.

\subsection{Text Summarization}
\label{sec:text_summ}

While FinGPT demonstrated strong performance across several financial NLP tasks, its ability to perform abstractive text summarization was notably limited. This was assessed using the \texttt{flare-ECTSum} dataset, which consists of financial event summaries derived from economic news sources. Despite several attempts, FinGPT failed to generate coherent or informative summaries, often producing generic or fragmented text unrelated to the input.

This shortcoming prompted a technical investigation into the architectural limitations of FinGPT’s underlying model—LLaMA—particularly in the context of tasks requiring long-range dependency modeling and bidirectional context understanding.

\subsubsection{Limitations of Decoder-Only Architectures for Summarization}

FinGPT is built on the LLaMA architecture, a decoder-only causal language model (CLM) trained using the standard autoregressive objective:

\[
P(x_1, x_2, \ldots, x_n) = \prod_{t=1}^{n} P(x_t \mid x_1, x_2, \ldots, x_{t-1})
\]

This unidirectional generation paradigm is highly effective for tasks such as next-token prediction, general text generation, and constrained question answering. However, it introduces inherent limitations when applied to abstractive summarization tasks, which require a global understanding of the entire input before generating a condensed version.

In contrast, encoder-decoder models like T5 and BART are designed specifically for conditional sequence generation. These models learn the probability of generating a summary sequence $y = (y_1, \ldots, y_m)$ given an input sequence $x = (x_1, \ldots, x_n)$ as follows:

\[
P(y_1, y_2, \ldots, y_m \mid x_1, x_2, \ldots, x_n)
\]

In this formulation:
\begin{itemize}
    \item The encoder processes the full input $x$ to create a context-rich representation.
    \item The decoder generates the output $y$ conditioned on this representation, allowing it to consider the entire document structure and semantics.
\end{itemize}

This bidirectional encoding mechanism enables encoder-decoder models to excel at summarization, particularly in financial texts where salient information may be spread across multiple clauses or sentences.

\subsubsection{Implication for FinGPT}

Given its decoder-only design, FinGPT lacks the full-context representation necessary for accurate and concise summarization. Additionally, financial text often contains complex structures, nested events, and high information density—all of which are poorly captured when the model can only attend to past tokens. This architectural constraint explains the model’s poor performance on the \texttt{flare-ECTSum} dataset.

Future work could involve incorporating retrieval-augmented generation (RAG) or transitioning to hybrid or encoder-decoder models to improve summarization performance in domain-specific LLMs such as FinGPT.

\subsection{Summary Table}
\label{sec:summary_table}

Table~\ref{tab:benchmark_comparison} provides a comprehensive summary of FinGPT's performance across six key financial NLP tasks, benchmarked against FinMA 7B, GPT-4, human performance (where available), and traditional baselines.

FinGPT demonstrates strong performance in sentiment analysis and text classification, with F1-scores rivaling or surpassing GPT-4 in certain cases. In named entity recognition, the model achieves moderate accuracy, though it trails GPT-4, indicating room for improvement in structured output tasks.

In contrast, FinGPT shows significant limitations in tasks that require numerical reasoning and deep context understanding—most notably in financial question answering (QA), where performance lags well behind GPT-4 and human baselines. Summarization remains the most challenging task, as expected for decoder-only architectures like LLaMA.

For stock movement prediction, FinGPT performs moderately across three financial datasets. While not state-of-the-art, these results highlight its generalization ability in high-volatility financial forecasting scenarios.

These findings underscore the importance of domain-specific tuning, model architecture, and task complexity in evaluating large language models for finance.

\begin{table}[h!]
\footnotesize
\centering
\caption{Comparative Performance (\%) Across Tasks for FinGPT, FinMA 7B, GPT-4, Human, and Baseline Benchmarks}
\label{tab:benchmark_comparison}
\setlength{\tabcolsep}{3pt}
\begin{tabular}{l|l|c|c|c|c|c}
\toprule
\textbf{Task} & \textbf{Dataset} & \textbf{FinGPT} & \textbf{FinMA 7B} & \textbf{GPT-4} & \textbf{Human} & \textbf{Baseline} \\
\midrule
SA  & flare-FPB     & 87.62 (F1) & 87.00 (F1) & 86.00 (F1) & --     & 82.00 (F1) \\
    & FiQA-SA       & 95.80 (F1) & 79.00 (F1) & 88.00 (F1) & --     & -- \\
\cmidrule{1-7}
TC  & Headline      & 95.50 (F1) & 97.00 (F1) & 86.00 (Avg. F1) & -- & 94.20 (F1) \\
\cmidrule{1-7}
NER & FIN           & 69.76 (Entity.F1) & 69.00 (Entity.F1) & 83.00 (Entity.F1) & -- & 67.30 (F1) \\
\cmidrule{1-7}
QA  & ConvFinQA     & 28.47 (EM) & 20.00 (EM) & 76.00 (EM) & 89.00 (EM) & -- \\
    & FinQA         & 3.80 (EM)  & 4.00 (EM)  & 69.00 (EM) & 91.00 (EM) & -- \\
\cmidrule{1-7}
SMP & CIKM18        & 45.00 (F1) & 53.00 (F1) & 57.00 (Acc) & -- & -- \\
    & StockNet      & 48.22 (F1) & 56.00 (F1) & 52.00 (Acc) & -- & -- \\
    & BigData22     & 52.74 (F1) & 49.00 (F1) & 54.00 (Acc) & -- & -- \\
\cmidrule{1-7}
TS  & ECTSum        & --         & 8.00 (ROUGE-1) & 30.00 (ROUGE-1) & -- & -- \\
\bottomrule
\end{tabular}
\end{table}

\begin{figure}[h!]
    \centering
    \begin{subfigure}[b]{0.45\linewidth}
        \includegraphics[width=\linewidth]{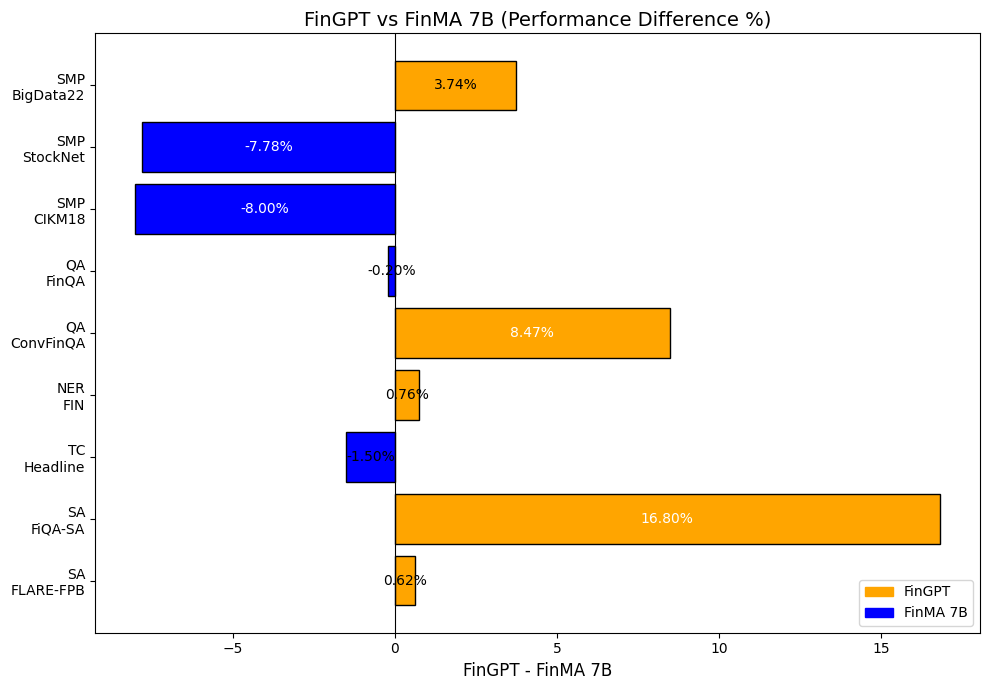}
        \caption{FinGPT vs FinMA 7B}
    \end{subfigure}
    \hfill
    \begin{subfigure}[b]{0.45\linewidth}
        \includegraphics[width=\linewidth]{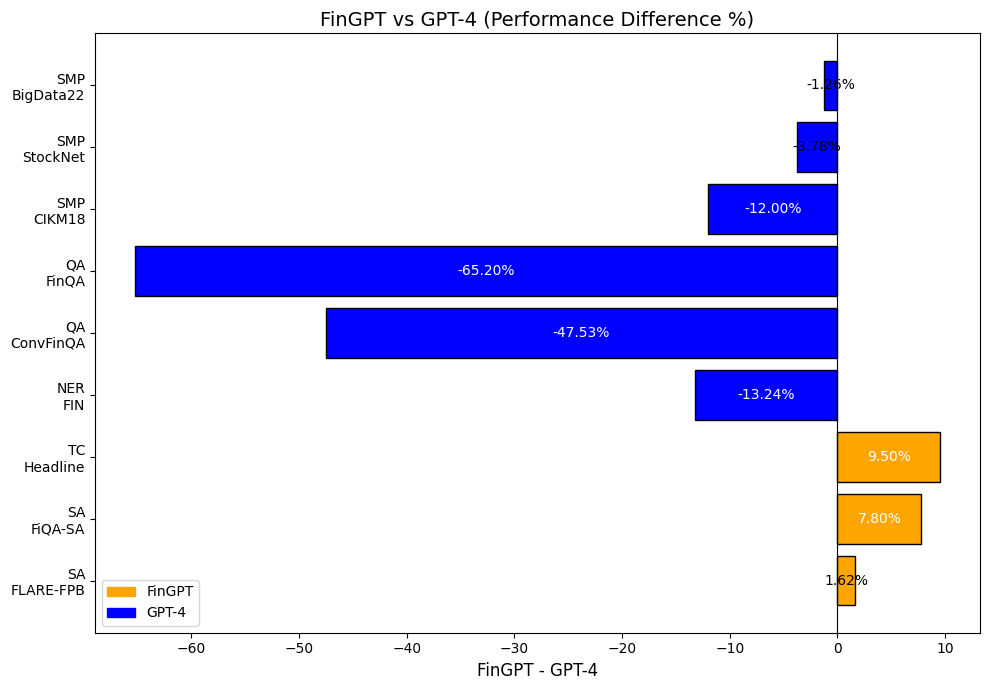}
        \caption{FinGPT vs GPT-4}
    \end{subfigure}
\end{figure}

\begin{figure}[h!]
    \centering
    \begin{subfigure}[b]{0.42\linewidth}
        \includegraphics[width=\linewidth]{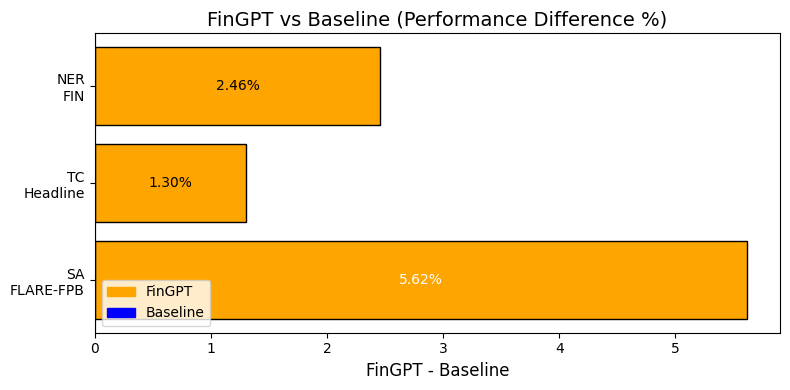}
        \caption{FinGPT vs Baseline Acc}
    \end{subfigure}
    \hfill
    \begin{subfigure}[b]{0.42\linewidth}
        \includegraphics[width=\linewidth]{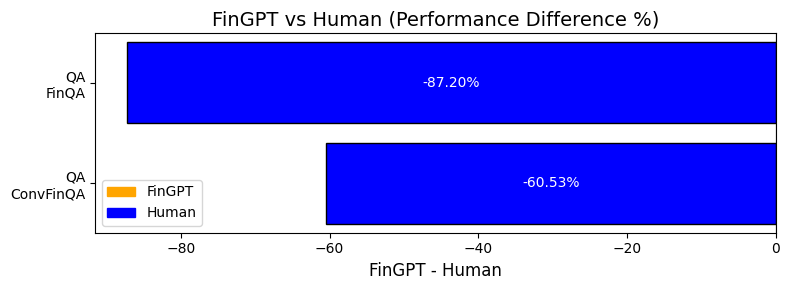}
        \caption{FinGPT vs Human Acc}
    \end{subfigure}
\end{figure}

\newpage
\section{Results and discussions} 
\label{sec:results}

This chapter presents the empirical results of the evaluation of FinGPT in six key financial NLP tasks. The analysis highlights performance strengths and weaknesses compared to domain-specific models (e.g. FinMA 7B), general-purpose models (e.g. GPT-4), and baseline or human references. Each section includes quantitative metrics and corresponding interpretations.

\section{Sentiment Analysis}

\begin{table}[h!]
\centering
\caption{Accuracy Comparison on Financial Sentiment Classification}
\label{tab:sentiment-results}
\setlength{\tabcolsep}{6pt}
\scriptsize
\begin{tabular}{l|c|c|c|c|c}
\toprule
\textbf{Dataset} & \textbf{FinGPT (Acc)} & \textbf{FinMA 7B (F1)} & \textbf{GPT-4 (F1)} & \textbf{Human Acc} & \textbf{Baseline (F1)} \\
\midrule
FLARE-FPB       & 87.62\% & 87.00\% & 86.00\% & --   & 82.00\% \\
FLARE-FIQA-SA   & 95.74\% & 79.00\% & 88.00\% & --   & --      \\
\bottomrule
\end{tabular}
\end{table}

\begin{figure}[h!]
    \centering
    \includegraphics[width=0.8\linewidth]{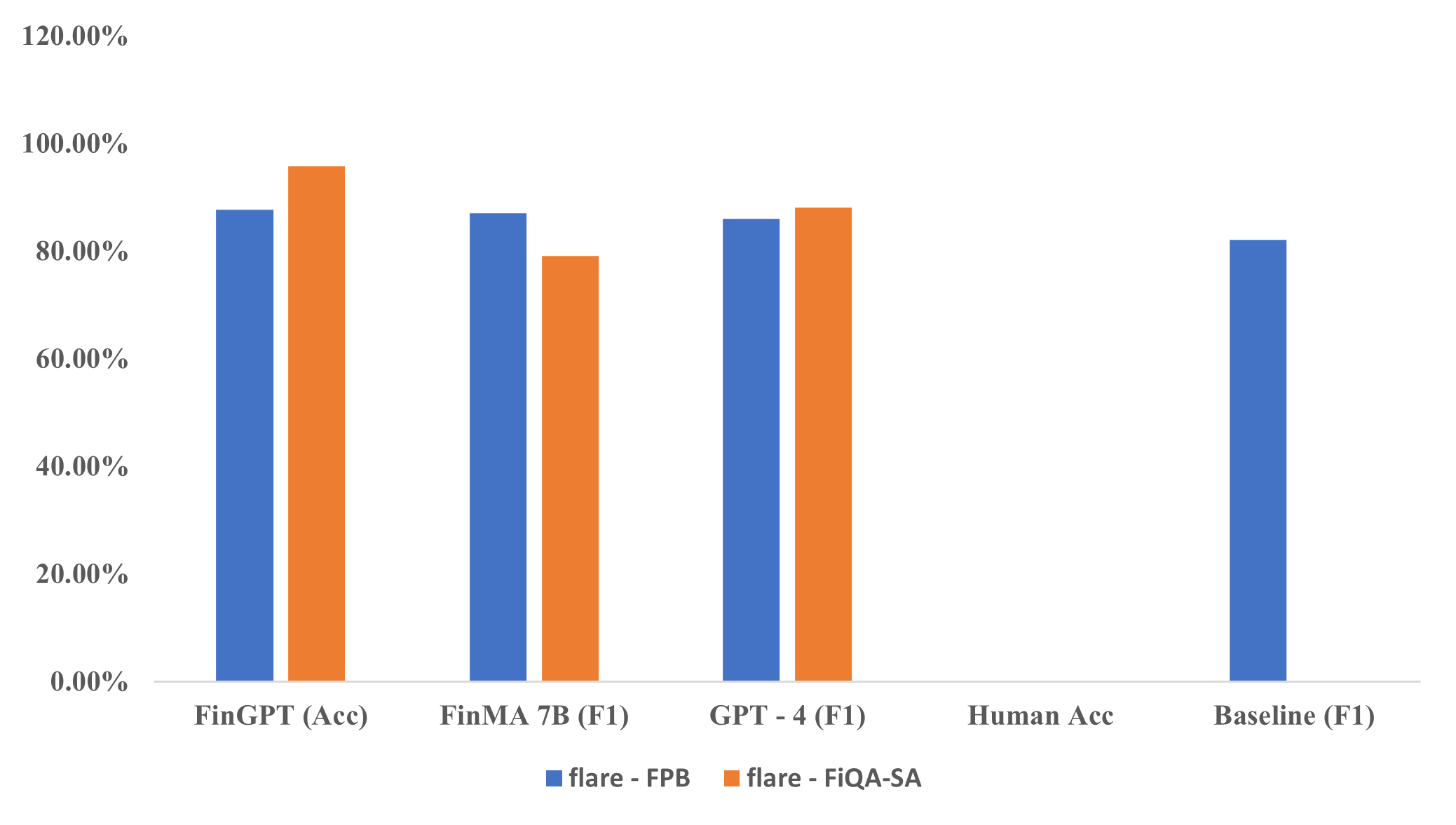}
    \caption{Performance Comparison on Financial Sentiment Datasets}
    \label{fig:sentiment-performance}
\end{figure}

\noindent
\textbf{Interpretation:} FinGPT demonstrated strong performance on both sentiment analysis datasets. On FLARE-FPB, it marginally outperformed GPT-4 and FinMA 7B, and achieved the highest accuracy on FLARE-FIQA-SA. These results suggest that FinGPT is highly effective at interpreting sentiment in financial texts, outperforming the baseline and benefiting from domain-specific alignment.

\section{Text Classification}
\begin{table}[h!]
\centering
\caption{Average F1 Score Comparison for Financial Headline Classification}
\label{tab:tc-results}
\setlength{\tabcolsep}{6pt}
\scriptsize
\begin{tabular}{l|c|c|c|c|c}
\toprule
\textbf{Model} & \textbf{FinGPT (Avg. F1)} & \textbf{FinMA 7B (F1)} & \textbf{GPT-4 (Avg. F1)} & \textbf{Human} & \textbf{Baseline (F1)} \\
\midrule
Headline Dataset & 95.50\% & 97.00\% & 86.00\% & -- & 94.20\% \\
\bottomrule
\end{tabular}
\end{table}
\newpage
\begin{figure}[h!]
    \centering
    \includegraphics[width=0.7\linewidth]{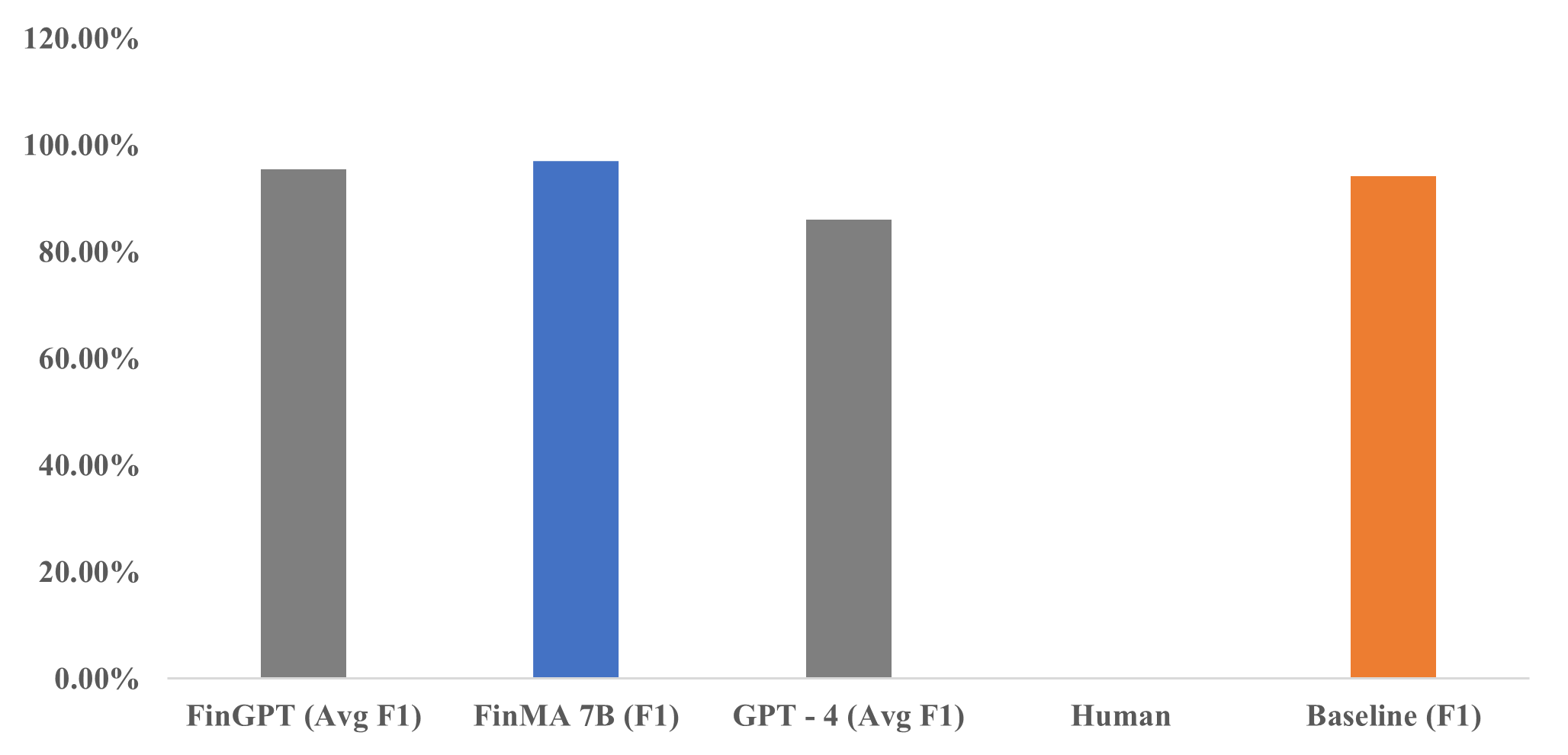}
    \caption{Performance on Headline-Based Text Classification}
    \label{fig:text-classification-performance}
\end{figure}

\noindent
\textbf{Interpretation:} On the headline classification task, FinGPT performed competitively, achieving an average F1 score of 95.5\%, slightly above the baseline and GPT-4. FinMA 7B attained the highest score, highlighting the potential advantage of more extensive fine-tuning. Nonetheless, FinGPT proves effective for financial text classification, reinforcing the utility of specialized language models in finance.

\section{Named Entity Recognition (NER)}

\begin{table}[h!]
\centering
\caption{Entity-Level F1 Comparison for Named Entity Recognition (NER)}
\label{tab:ner-results}
\setlength{\tabcolsep}{6pt}
\scriptsize
\begin{tabular}{l|c|c|c|c}
\toprule
\textbf{Model} & \textbf{FinGPT (Entity F1)} & \textbf{FinMA 7B (F1)} & \textbf{GPT-4 (Entity F1)} & \textbf{Baseline (F1)} \\
\midrule
FIN Dataset & 69.76\% & 69.00\% & 83.00\% & 67.30\% \\
\bottomrule
\end{tabular}
\end{table}

\begin{figure}[h!]
    \centering
    \includegraphics[width=0.8\linewidth]{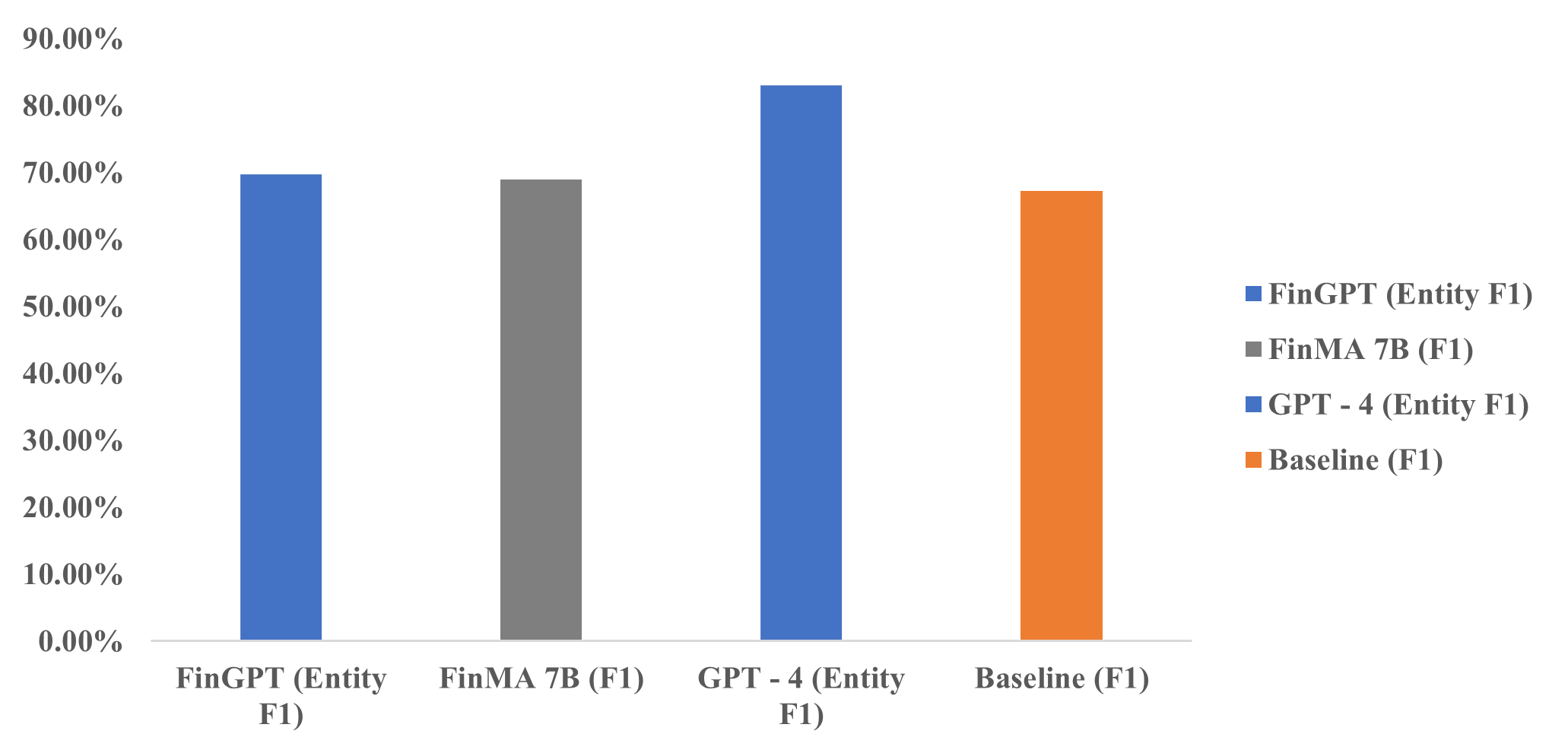}
    \caption{Comparison for Named Entity Recognition (NER)}
    \label{fig:ner-performance}
\end{figure}
\newpage
\noindent
\textbf{Interpretation:} FinGPT and FinMA 7B show comparable results, slightly outperforming the baseline. However, GPT-4 significantly surpasses both models, indicating a superior ability to extract and categorize financial entities. This underscores the challenge of complex entity recognition in finance and reveals areas where FinGPT can benefit from further refinement.

\section{Financial Question Answering}

\begin{table}[h!]
\centering
\caption{Exact Match (EM) Accuracy Comparison for Financial QA}
\label{tab:fqa_results}
\setlength{\tabcolsep}{6pt}
\scriptsize
\begin{tabular}{l|c|c|c|c}
\toprule
\textbf{Dataset} & \textbf{FinGPT (EM)} & \textbf{FinMA 7B (EM)} & \textbf{GPT-4 (EM)} & \textbf{Human (EM)} \\
\midrule
ConvFinQA     & 28.4\% & 20.0\% & 76.0\% & 89.0\% \\
FLARE-FinQA   & 3.8\%  & 4.0\%  & 69.0\% & 91.0\% \\
\bottomrule
\end{tabular}
\end{table}

\begin{figure}[h!]
    \centering
    \includegraphics[width=0.7\linewidth]{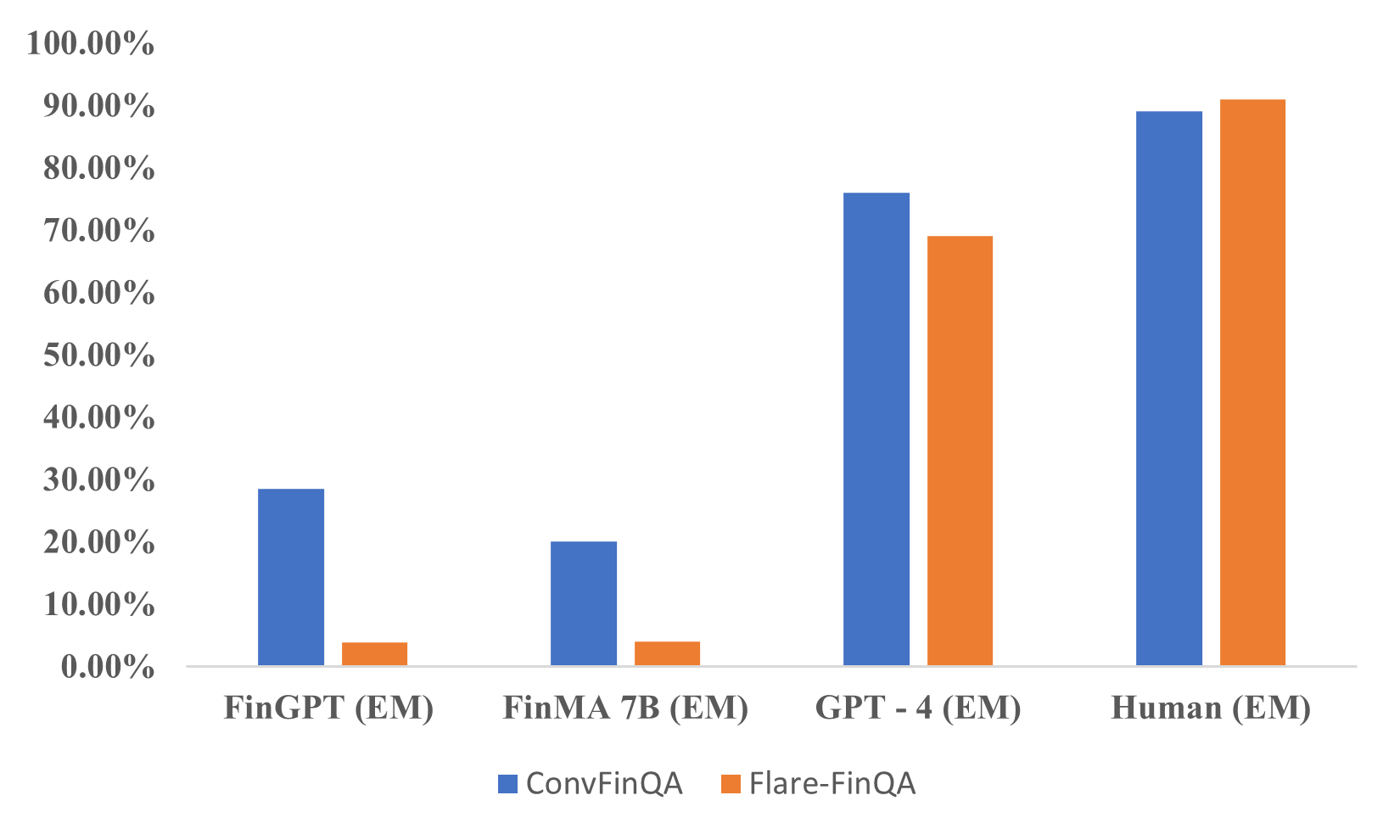}
    \caption{Comparison of Model Performance on Financial QA Datasets}
    \label{fig:qa-performance}
\end{figure}

\noindent
\textbf{Interpretation:} FinGPT and FinMA 7B struggle with financial question answering, particularly on FLARE-FinQA, which demands complex reasoning and precise numerical answers. The gap between these models and GPT-4 or human performance is substantial, indicating that existing domain-specific models require deeper reasoning capabilities and enhanced numerical understanding.

\section{Stock Movement Prediction (SMP)}

\begin{table}[h!]
\centering
\caption{Accuracy Comparison for Stock Movement Prediction}
\label{tab:f1-stocknet-cikm18}
\setlength{\tabcolsep}{6pt}
\scriptsize
\begin{tabular}{l|c|c|c|c}
\toprule
\textbf{Dataset} & \textbf{FinGPT (Acc)} & \textbf{FinMA 7B (Acc)} & \textbf{GPT-4 (Acc)} & \textbf{Human Acc} \\
\midrule
StockNet   & 48.47\% & 56.00\% & 52.00\% & -- \\
CIKM18     & 47.03\% & 53.00\% & 57.00\% & -- \\
BigData22  & 52.83\% & 49.00\% & 54.00\% & -- \\
\bottomrule
\end{tabular}
\end{table}
\newpage
\begin{figure}[h!]
    \centering
    \includegraphics[width=0.6\linewidth]{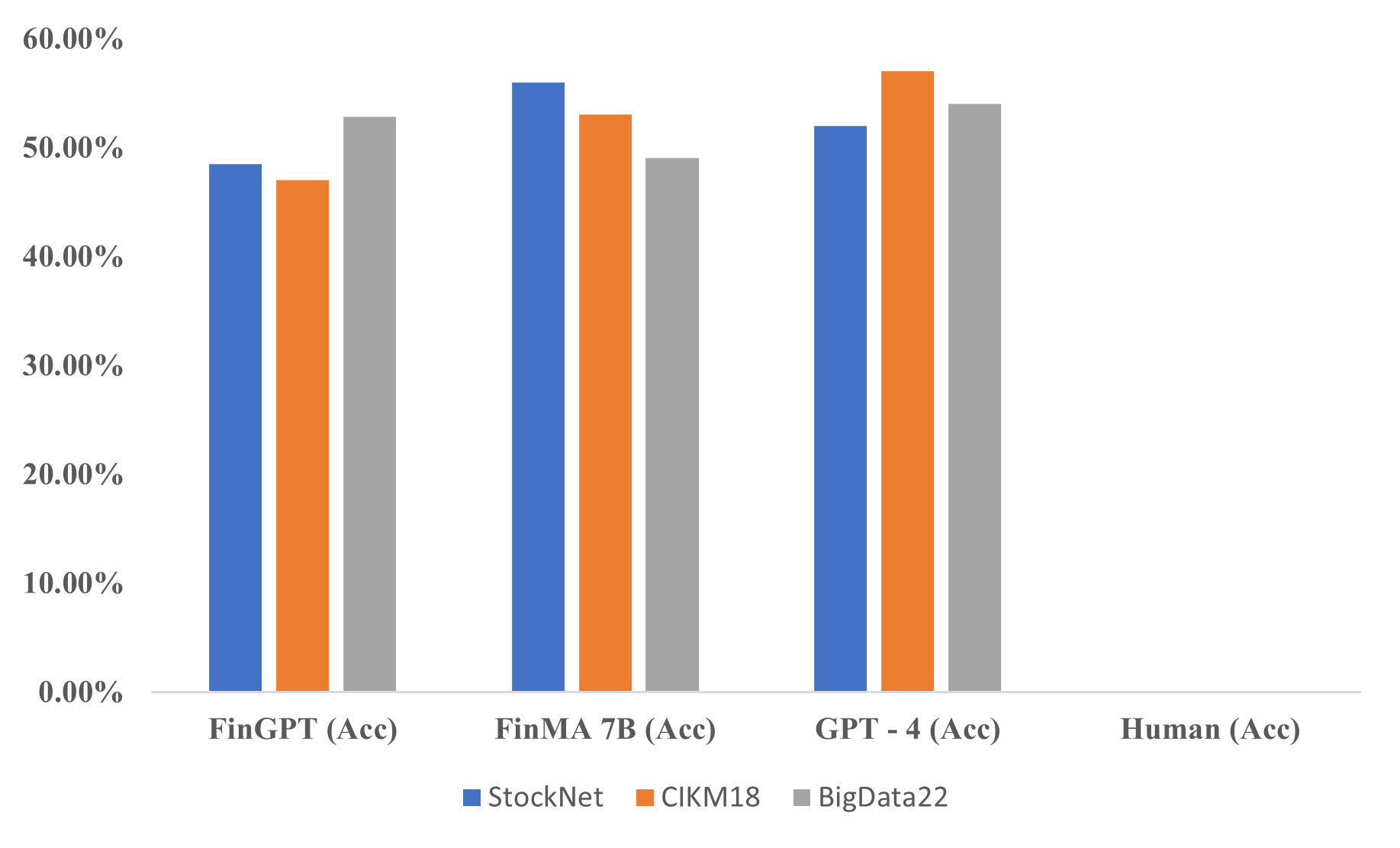}
    \caption{Stock Movement Prediction Accuracy Across Datasets}
    \label{fig:smp-datasets}
\end{figure}

\noindent
\textbf{Interpretation:} FinGPT demonstrates moderate accuracy across all stock movement datasets. While it generally trails behind GPT-4 and FinMA 7B, the results reflect the inherent difficulty of forecasting market trends using text data alone. These findings reinforce the challenge of financial prediction tasks, especially in the face of noise and volatility.

\subsection{Directional Sensitivity Analysis in Stock Movement Prediction}
\label{sec:smp_contribution}

To further assess FinGPT’s real-world applicability, we conducted a directional sensitivity analysis to evaluate whether it performs better in bullish or bearish market phases.

\subsubsection{CIKM18 Dataset}

\begin{figure}[h!]
    \centering
    \includegraphics[width=0.7\textwidth]{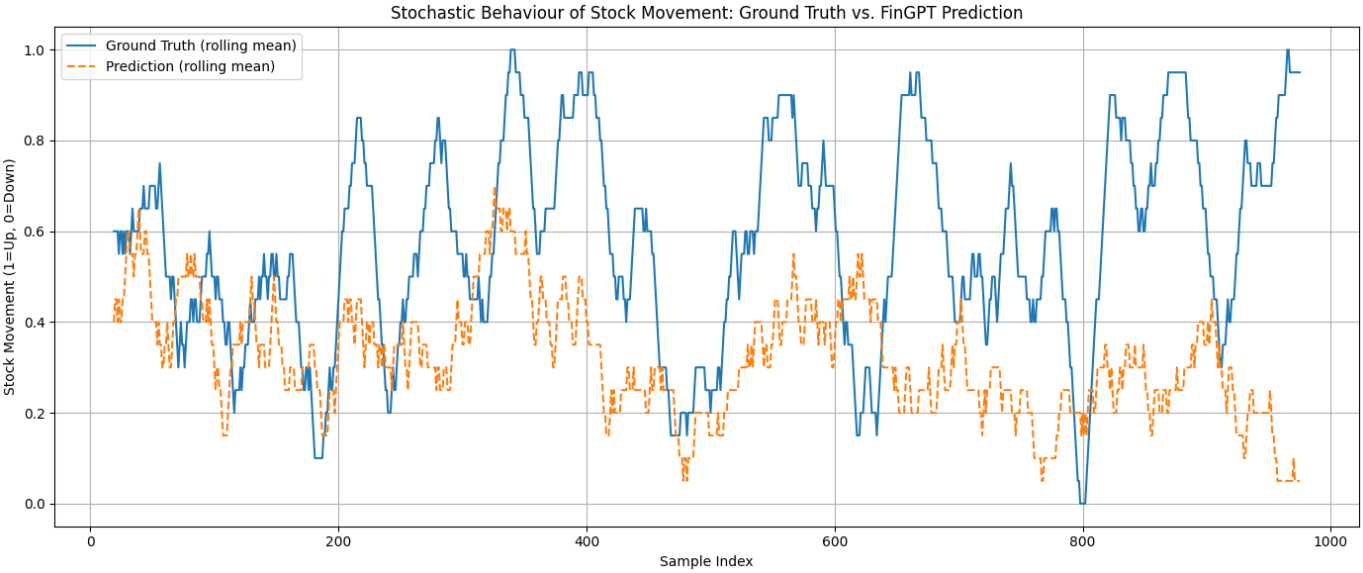} 
    \caption{Rolling Mean of FinGPT vs Ground Truth on CIKM18}
    \label{fig:smp_cikm18}
\end{figure}

\noindent
As shown in Figure~\ref{fig:smp_cikm18}, FinGPT aligns closely with upward trends but lags or flattens during market downturns. This suggests a directional bias toward bullish sentiment.

\subsubsection{Directional Bias Observation}

\begin{figure}[h!]
    \centering
    \includegraphics[width=0.7\textwidth]{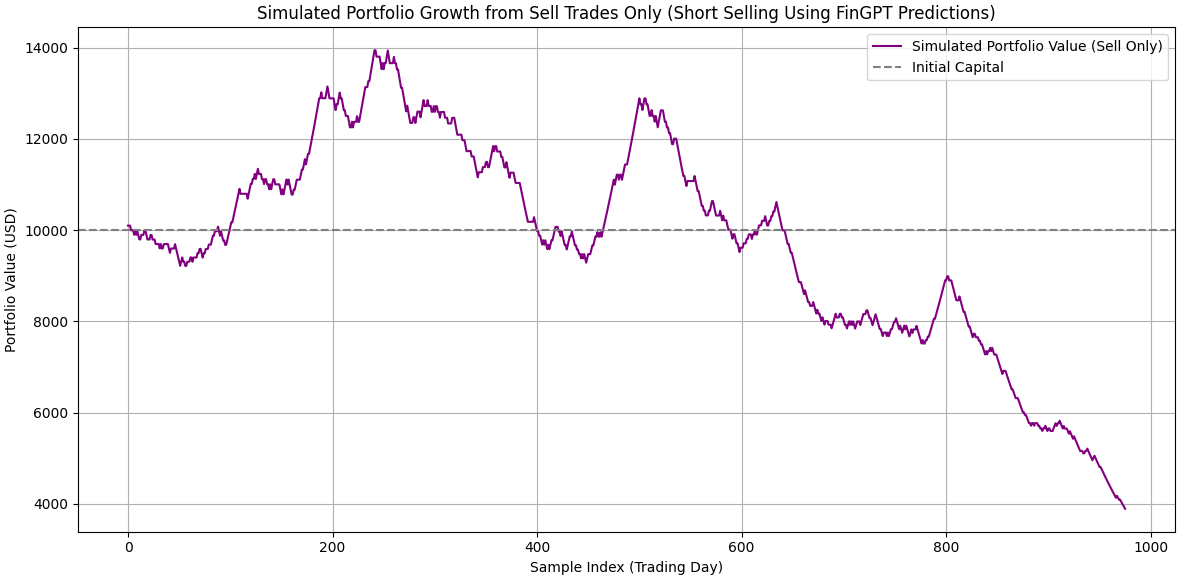}
    \caption{Portfolio Value (Bearish Trades) — CIKM18}
    \label{fig:smp_stocknet}
\end{figure}

\begin{figure}[h!]
    \centering
    \includegraphics[width=0.7\textwidth]{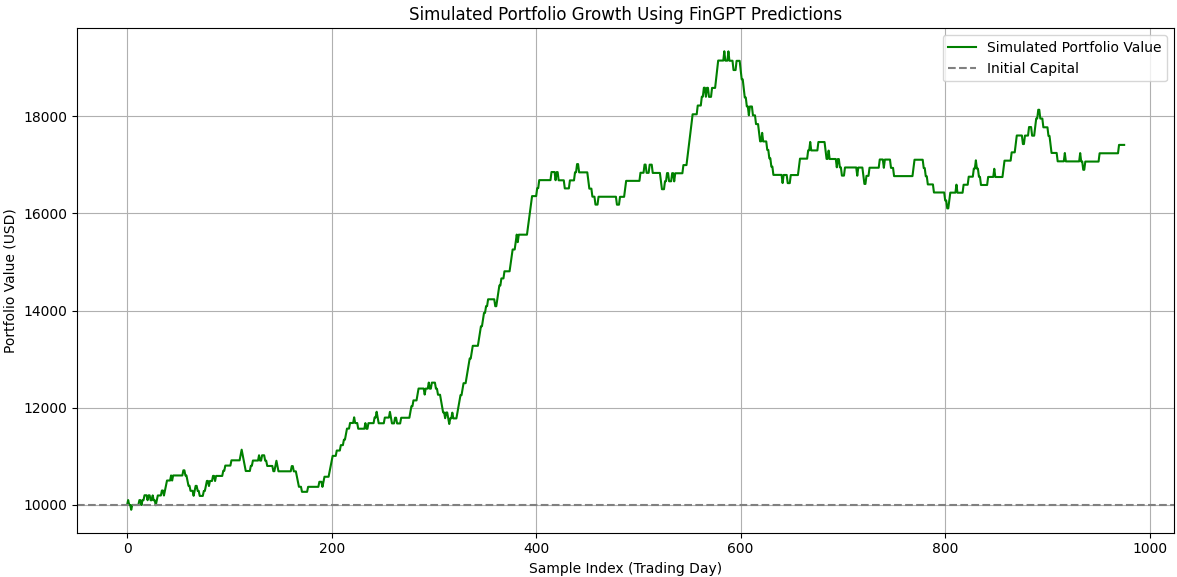}
    \caption{Portfolio Value (Bullish Trades) — CIKM18}
    \label{fig:smp_bigdata22}
\end{figure}

This bullish bias is consistent across datasets. Figures~\ref{fig:smp_stocknet} and \ref{fig:smp_bigdata22} display simulated portfolio performance, showing stronger returns when the model predicts buy signals versus sell signals.

\newpage
\subsubsection{BigData22 Dataset}

\begin{figure}[h!]
    \centering
    \includegraphics[width=0.7\textwidth]{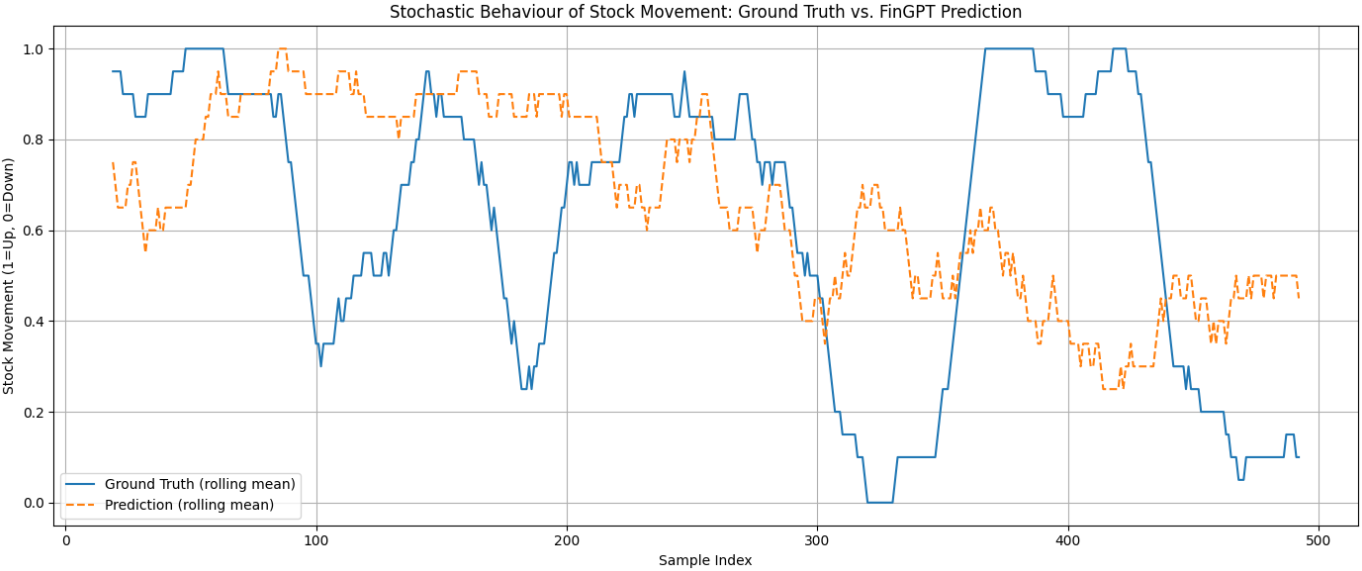}
    \caption{FinGPT vs Ground Truth (BigData22)}
    \label{fig:smp_bigdata22_roll}
\end{figure}

\noindent
Figure~\ref{fig:smp_bigdata22_roll} reaffirms FinGPT’s preference for bullish signal tracking. Its performance dips in response to sharp downturns but aligns more accurately during uptrends.

\begin{figure}[h!]
    \centering
    \includegraphics[width=0.7\textwidth]{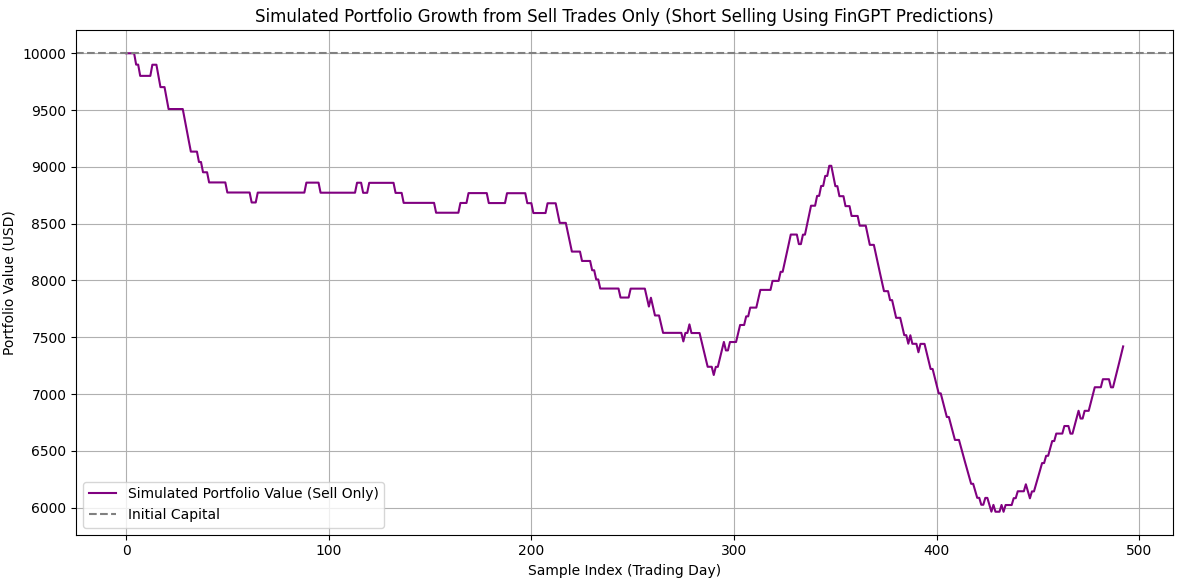}
    \caption{Portfolio Value from Bearish Trades (BigData22)}
    \label{fig:smp_bigdata22_bear}
\end{figure}

\begin{figure}[h!]
    \centering
    \includegraphics[width=0.7\textwidth]{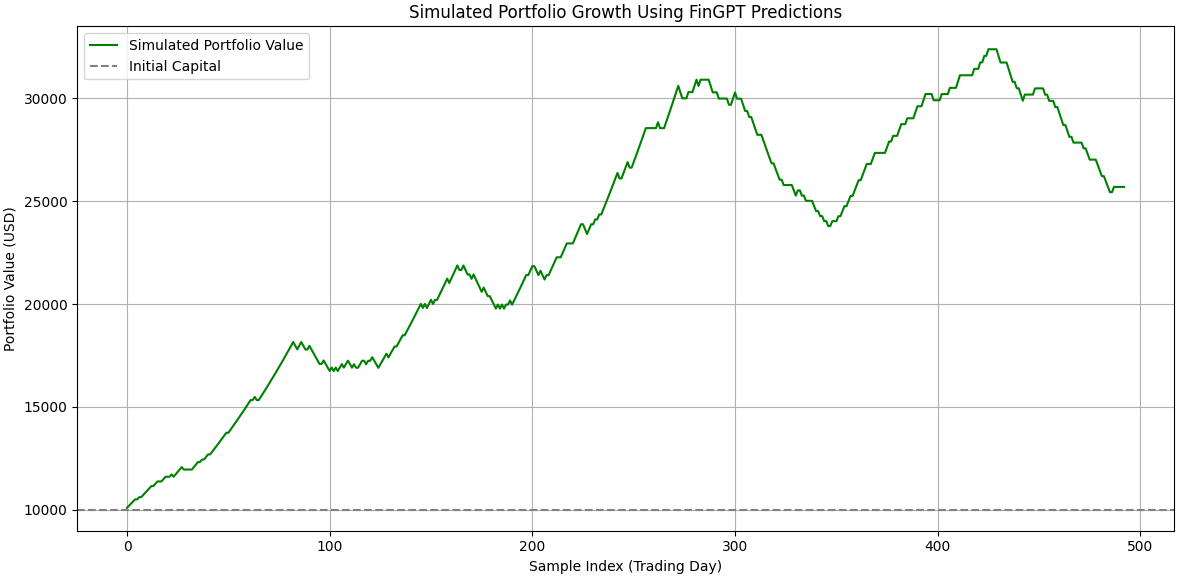}
    \caption{Portfolio Value from Bullish Trades (BigData22)}
    \label{fig:smp_bigdata22_bull}
\end{figure}

\newpage
\subsubsection{StockNet Dataset}

\begin{figure}[h!]
    \centering
    \includegraphics[width=0.7\textwidth]{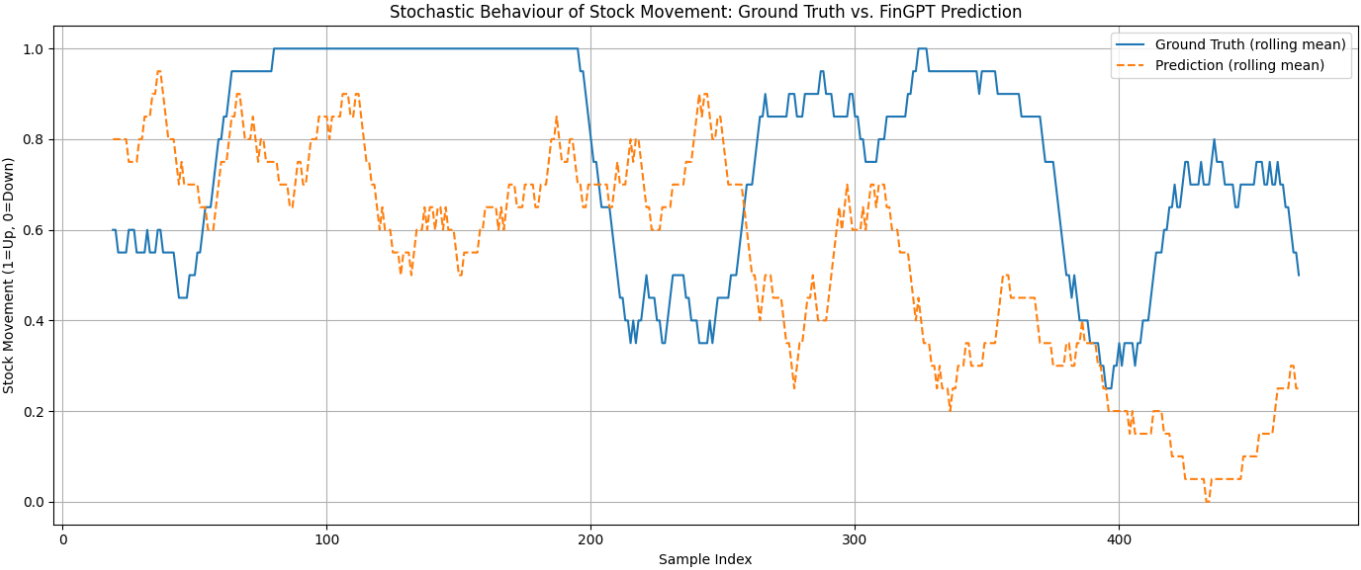}
    \caption{Smoothed Stock Movement on StockNet Dataset}
    \label{fig:stocknet_rolling}
\end{figure}

\paragraph{Trading Simulation:} Two strategies were tested:
\begin{itemize}
    \item \textbf{Short-only:} Sell when FinGPT predicts market decline.
    \item \textbf{Long-only:} Buy when FinGPT predicts a rise.
\end{itemize}

\begin{figure}[h!]
    \centering
    \includegraphics[width=0.7\textwidth]{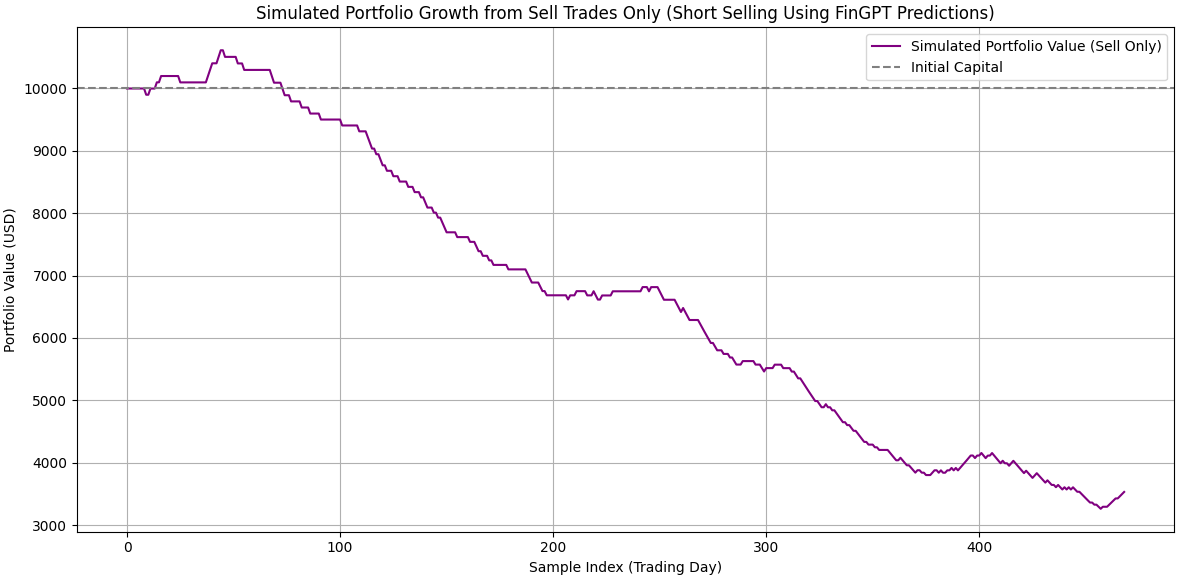}
    \caption{Short-Only Strategy Portfolio — StockNet}
    \label{fig:stocknet_bearish}
\end{figure}

\begin{figure}[h!]
    \centering
    \includegraphics[width=0.7\textwidth]{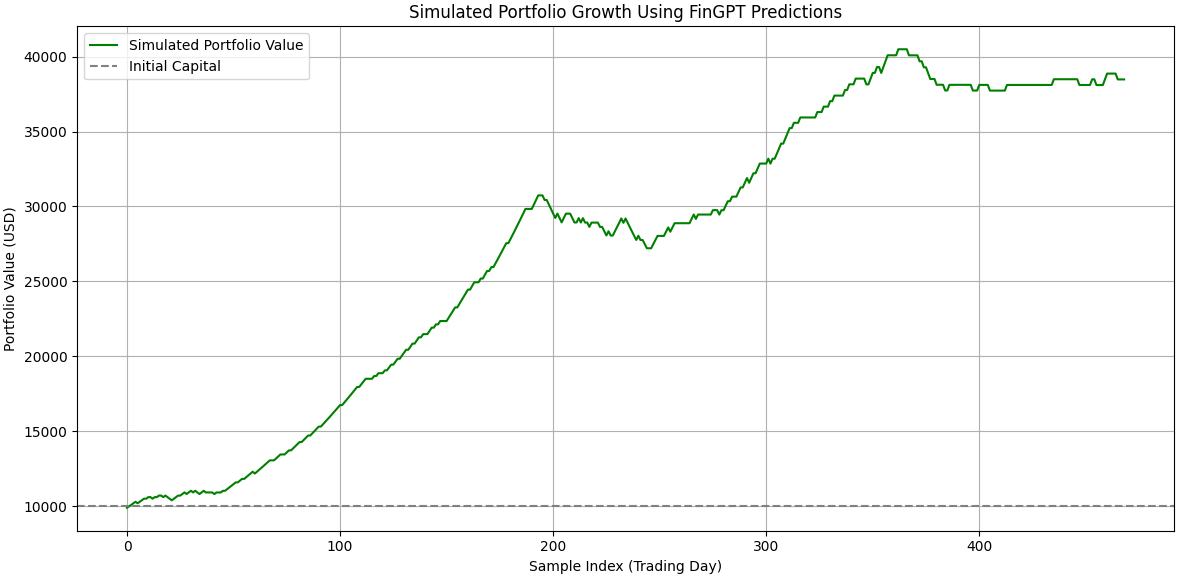}
    \caption{Long-Only Strategy Portfolio — StockNet}
    \label{fig:stocknet_bullish}
\end{figure}

\paragraph{Insights:} FinGPT’s predictive alignment with bullish markets leads to positive trading returns in upward trends, but its underperformance in bearish contexts highlights a key limitation. This directional imbalance is important for practical deployment in trading systems and suggests a need for improved calibration or training on diverse market conditions.

\section{Conclusion}
This research evaluated FinGPT performance on six key financial NLP tasks. The model showed strong results in classification tasks like sentiment analysis and headline classification, often matching or exceeding general models. However, it struggled with reasoning-heavy tasks like question answering and summarization. Benchmarking against GPT-4, FinMA 7B, and human performance provided valuable context. While FinGPT is promising for finance-specific NLP, it is not yet a full substitute for more advanced or general-purpose models. The findings underscore the importance of model architecture and computational resources in achieving reliable performance. This work lays a solid foundation for future improvements in domain-specific language models and highlights FinGPT’s potential in targeted financial applications.

\section{Key Findings}
\label{sec:key_findings}

\subsection{Financial Question Answering (QA)}

FinGPT demonstrated significant limitations in answering multi-step, numerically grounded financial questions, particularly when evaluated on datasets such as ConvFinQA and FLARE-FinQA. The Exact Match (EM) scores (3.8\%–28.4\%) fell well below human-level (89–91\%) and GPT-4 (69–76\%) benchmarks. These results indicate that FinGPT, while domain-aligned, lacks the deep reasoning and arithmetic capabilities required to handle quantitative financial queries. This suggests a key limitation in FinGPT’s autoregressive architecture and its training setup. Unlike general-purpose models augmented with retrieval or external tools, FinGPT operates with limited reasoning depth, revealing a gap between task-specific expectations and model capacity. Future directions may involve augmenting FinGPT with symbolic reasoning, external calculators, or chain-of-thought prompting strategies.

\subsection{Stock Movement Prediction (SMP)}

The SMP task remains inherently challenging due to the stochastic nature of financial markets. In our experiments, FinGPT achieved modest accuracy scores across the CIKM18, StockNet, and BigData22 datasets (ranging from 47.0\% to 52.8\%). These results place it slightly behind FinMA 7B and GPT-4, but still ahead of traditional baselines in some configurations. What sets this task apart is the complexity of associating unstructured text with structured stock price movements—a task highly sensitive to both news content and market context. Despite lacking prior strong baselines, FinGPT offers a credible first benchmark, particularly when extended through our new directional analysis framework.


To expand the interpretability of FinGPT’s predictions beyond standard classification accuracy, we introduced a novel directional sensitivity analysis. This bi-directional performance evaluation measures how well the model detects and reacts to bullish versus bearish signals over time.

Across all three datasets, FinGPT exhibited a consistent bias toward bullish (upward) trends. While this bias enabled strong predictive alignment in rising markets, it also caused performance degradation during market downturns. Simulated trading portfolios built on FinGPT’s buy/sell signals revealed clear asymmetries: long-only strategies (buying on bullish predictions) consistently outperformed short-only strategies. This analysis offers two major insights:

\begin{enumerate}
    \item \textbf{Model behavior is asymmetrical:} FinGPT is more attuned to upward sentiment, possibly due to training data imbalances or architectural bias in autoregressive decoding.
    
    \item \textbf{Implications for real-world deployment:} In practice, FinGPT may be more reliable during growth phases than recessions. This insight is critical for designing trading strategies or decision-support systems that integrate model predictions.
\end{enumerate}

Together, this bidirectional analysis introduces a valuable evaluation perspective for financial forecasting models, moving beyond flat metrics to behavioral insights. It also sets a precedent for future FinLLMs to incorporate risk-aware performance evaluations.



\newpage

\bibliography{biblio_jmmbiblio_jmm}
\bibliographystyle{plain}
\end{document}